\pdfoutput=1
\documentclass[conference]{IEEEtran}
\usepackage{amssymb}
\usepackage{amsmath}
\usepackage{algorithm}
\usepackage{algpseudocode}
\usepackage{graphicx}
\usepackage{tikz}
\usepackage{arydshln}
\usetikzlibrary{positioning,shapes,arrows}
\usepackage{mdframed}
\usepackage{xcolor}

\begin{document}
%
\title{Enabling Basic Normative HRI in a Cognitive Robotic
  Architecture}

\author{\IEEEauthorblockN{Vasanth Sarathy, Jason R. Wilson, Thomas Arnold and Matthias Scheutz}
\IEEEauthorblockA{Human-Robot Interaction Laboratory, Tufts University\\
Medford, MA 02155, USA\\
Email: \{vsarathy,wilson\}@cs.tufts.edu, \{thomas.arnold,matthias.scheutz\}@tufts.edu}}

\maketitle



%
\IEEEpeerreviewmaketitle

\section{Introduction}

Collaborative human activities are grounded in {\em social and moral
  norms}, which humans consciously and subconsciously use to guide and
constrain their behavior: they undergird human societies by prescribing what is obligatory,
permitted, prohibited, and optional \cite{brown1991human}.  In doing so, they enable
effective collaboration and prevent emotional and physical harm.

Consider a restaurant kitchen where cooks and assistants perform tasks
such as passing knives and cutting vegetables. When handing over a
knife to the chef, assistants do so in a way that does not look like
they are about to stab the chef. Not only will they orient the knife
in the right way, but they should take care not to approach the chef
menacingly and without prior warning, while the chef has their back to
them.  The underlying normative principle could be roughly stated as a
rule: ``If you need to hand over a potentially dangerous object with a
sharp blade, do not point it with the blade at the other person, but
rather grasp it carefully by the blade and hand it over with the bland
side or handle facing the other person''.  The tacit understanding
among the kitchen staff is that everyone will abide by this
principle, thus enabling safe exchanges of knives and other
potentially dangerous objects.  Failing to follow the rule will likely
result in blame and reprimand from the chef, which then has to be
addressed either by apologizing or by offering an explanation as to
why the rule violation was justified \cite{malle2014moral}.

Clearly, social and moral norms play an important functional role in
the human cognitive architecture: they are at work in perception to
detect morally charged contexts and norm violations, they are employed
during decision-making and behavior execution, and they are referred
to in communications about normative behavior and norm violations.  In
other words, normative processing is deeply integrated into the human
cognitive system and affects virtually every aspect of the
architecture (from perception, to reasoning, to action, to
communication).  Hence, this type of norm-based processing is also
critical for robots in many human-robot interaction scenarios (e.g.,
when helping elderly and disabled persons in assisted living
facilities, or assisting humans in assembly tasks in factories or even
the space station).  Human beings expect their interactants, including
intelligent robots, to follow social and moral norms, and
disappointing those expectations will lead to impoverished
interactions at best, but can lead to emotional and physical harm in
the worst cases.  


In this position paper, we will briefly describe how several
components in an integrated cognitive architecture can be used to
implement processes that are required for normative human-robot
interactions, especially in collaborative tasks where actions and
situations could potentially be perceived as threatening and thus need
a change in course of action to mitigate the perceived threats.  We
will focus on {\em affordance-based reasoning} to infer complex
affordance relationships between an agent and objects in the
environment, and {\em analogical reasoning} to decide the
appropriateness of the action plan by comparing against other past
situations.

\section{Background and Related Work}

  
 Many in the HRI field have recognized that ethics will have to inform competent robot behavior in the social sphere. Various ethical theories and combinations thereof have been proposed, most commonly weighing utilitarian approaches against deontic frameworks (e.g. obligated or forbidden actions), \cite{abney2012robotics}. Recent work in cognitive science on human moral reasoning, however, has recently yielded insights into intricate relationships between moral norms, emotions, theory of mind, and blame \cite{monroe2012morality}, \cite{greene2002and}. While autonomous social robots need not, indeed should not, recapitulate models or features of embodied human cognition for the sake of sheer similarity (e.g. reproducing ``aggression" that clouds moral principle), it is clear that to interact competently in social space such systems will incorporate adept perspective taking and reason giving for their actions \cite{scheutz2015towards}. Moreover, in their dealings with people robots will also be dealing with inanimate objects (from tools to keepsakes), which can be especially charged morally when involved with collaborative tasks or other social interactions: HRI cannot ignore object affordance in its scenarios and design for moral performance.
 
In terms of cognitive architecture, calls for robust moral reasoning have been building in scope and force as roles for AI and robotics, from self-driving cars and autonomous weapons systems to domestic and healthcare roles, have accelerated entry into the social sphere \cite{bello2013build}. Relatively little work on cognitive architecture has directly tackled social and moral norms, though there are some initial modeling efforts to meet that challenge \cite{sun2013moral}. MoralDM, for example, as part of the Companions architecture, base moral decision-making on analogies with cultural narratives 
\cite{dehghani2008integrated} or generalizations of stories \cite{blass2015moral} to determine the appropriate action. Recognizing how thoroughly moral norms will shape expectations and evaluations of social robots, we situate moral reasoning as an integral feature of the DIARC architecture  \cite{scheutzetal07autrobot}. 


\section{Components for Normative HRI}


Various architectural capabilities are required in cognitive robotic architectures for robots to become morally competent \cite{malle2014moral}. Here we focus on three key functionalities: (1) affordance inference, (2) analogical reasoning, and (3) action selection. For example, when handing over a knife, robots must pass the knife in a socially acceptable way (affordance perception), while evaluating whether the situation as a whole is socially appropriate compared with similar situations (analogical reasoning), and mitigating perceived threats by choosing alternative actions (action selection).

In prior work we have developed a computational representation and framework to reason about affordances more generally (i.e., physical, functional, aesthetic, social and ethical affordances) \cite{Sarathy2015}. We have also implemented a structure-mapping engine for analogical reasoning and have the ability to compare and score
situations for similarity in structure. Here we supplement this work with an action selection engine to reason about social and moral perceptions and select
mitigating actions.



We propose implementing these key functionalities by means of components integrated into the existing robotic DIARC architecture, which comprises components for perceptual and visual processing, navigation, action planning, and natural language processing. DIARC has been used extensively for human-robot interaction in natural language \cite{scheutzetal07autrobot}. Next, we will discuss each of these functionalities along with the architectural components needed to enact them  (Figure 1).

\begin{figure}[tb!]
\centering
\includegraphics[scale=2.6]{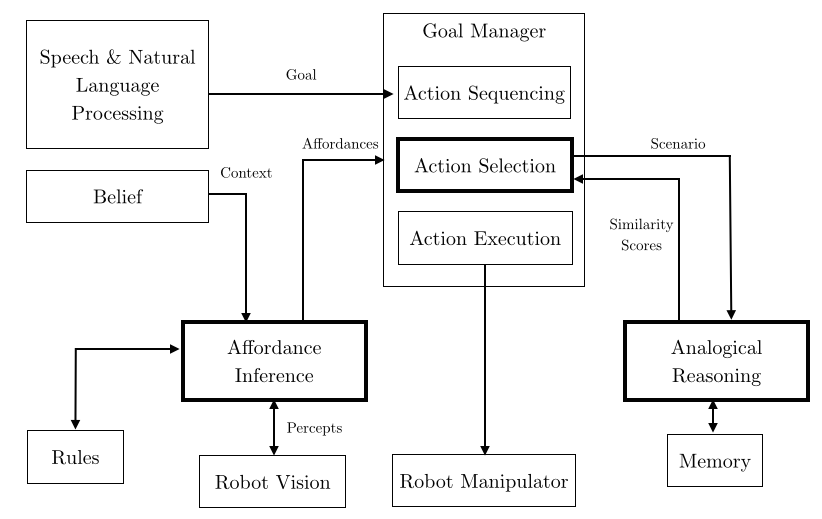}
\caption{High-level Schematic of DIARC showing the three relevant components together with some of their connections.}
\label{figurelabel}
\vspace{-3mm}
\end{figure}

\subsection{Goal Manager (GM)}

The Goal Manager (GM) is responsible for accepting a goal and assembling an action script to satisfy this goal and manages the execution of the script. The GM performs these functions in conjunction with the Affordance Inference Component and the Analogical Reasoning Component, which we will discuss in more detail below. 

\subsection{Cognitive Affordance Inference}

We have developed a formal rules-based logical representational format
and inference algorithm for cognitive affordances, in which an
object's affordance ($A$) and its perceived features ($F$) depend on
the context ($C$). The perceived features ($F$) include color, shape, texture, relational information, and general
information obtained from the vision (or other sensory systems)
pipeline that an agent may interpret. 
The context is representative of the agent's beliefs, goals,
desires, and intentions, along with certain other abstract constructs
in the agent's narrative situation. 
The object's affordance ($A$) represents the types of action possibilities that
might be available to the robot at any given moment in time.

We use Dempster-Shafer (DS) theory for inferring affordance ($A$) from
object features ($F$) in contexts ($C$) \cite{Shafer1976}. DS theory is an uncertainty
processing framework often interpreted as a generalization of the
Bayesian framework.

Our cognitive affordance model also consists of a
set of affordance rules ($R$) taking the form $r: \equiv f \land c \implies_{[\alpha,\beta]} a$ with $f\in F$, $c\in C$, $a\in A$, $r\in R$,
$[\alpha,\beta]\subseteq [0,1]$.  Here, the confidence interval
$[\alpha,\beta]$ is intended to capture the uncertainty associated
with the affordance rule $r$ such that if $\alpha=\beta=1$ the rule is
logically true, while $\alpha=0$ and $\beta=1$ assign maximum
uncertainty to the rule.  Rules can then be applied for a given
feature percept $f$ in given context $c$ to obtain the implied
affordance $a$ under uncertainty about $f$, $c$, and the extent to
which they imply the presence of $a$.

%

These types of rules are very versatile, and we can employ DS-theoretic modus ponens to make uncertain deductive and abductive inferences.  We have started to integrate this functionality into the DIARC architecture by means of a special affordance inference component in conjunction with the existing visual perception components, which allows us to incorporate cognitive affordance inference into the visual perception pipeline.

\subsection{Analogical Reasoning}

We use analogical reasoning to identify applicable actions that are
consistent with the surrounding context. The process proceeds as follows.  
Given an encoding of the situation we make a series of analogical comparisons with other
situations.  We use the Structure Mapping Engine (SME) \cite{falkenhainer1989structure}
to perform each comparison. The other situations are stored in memory and originate from prior experiences, instruction, observation, and demonstration.  Each successful
analogical comparison yields a similarity score and a set of
candidate inferences. Comparing the similarity scores of each
comparison indicates which situations are most
analogous to the current situation.  The candidate inferences
represent information known in the other situation that
structurally fits with the new situation.  Since there is no semantic
verification of this information, a follow-on step is necessary to
check each candidate inference and determine whether it can be true in
the current situation.  Included amongst the candidate inferences
may be the action for the robot to take or the perceived intent of the
action in a given context.

\section{Proof-of-Concept Example}

DIARC aims for ``natural, human-life" human-robot interaction through which a robot can deliver goal-oriented, socially appropriate behavior from exchanges in ordinary settings through natural language. To illustrate how the three components discussed above can contribute to that effort within DIARC, let us consider a robotic agent who is helping human beings at home in their kitchen. One of them asks the robot, ``Can you bring me something to cut a tomato?" The speech and natural language systems within DIARC can parse and interpret this instruction, submitting a goal to the GM (e.g., 
$possess(human, cutwith(tomato))$). The GM will resolve this goal into a high-level action script with three sequenced parts: $find$, $pick up$, and $bring$.    

\textbf{Find Object.}
Once the GM has resolved the larger goal into a hierarchical action script, each step of the action script is then further resolved into more primitive actions. The step of ``Find Object" is further resolved by turning to the Affordance Inference Component in the architecture. The Affordance Inference Component interacts with a set of affordance rules (which include physical, functional, and social
rules) stored in memory, where each rule associates an affordance
with certain perceptual features and certain contextual elements. In the kitchen-helper example, consider rule $r^{1}$ with an uncertainty interval $[0.8,1]$:

\begin{flushleft}
$r^1_{[0.8, 1]} := hasSharpEdge(O) \land domain(X,kitchen) \implies$ \\
	$cutWith(X,O)$\\
\end{flushleft}

The Affordance Inference Component receives contextual information (e.g., it is in the kitchen working as a helper) from a Belief component (tasked with resolving agent beliefs and intentions) and from the GM. It also interacts with the robot's vision component and directs a visual search to look for objects in the environment that satisfy the $hasSharpEdge(O)$ perceptual predicate. 

The Affordance Inference Component then applies perceptual and contextual information 
(along with accompanying uncertainty masses, $m$) to determine the affordance implied by the rule, as follows:

\begin{flushleft}
$r^1_{[0.8, 1]} (m_r = 0.8) := $\\
$hasSharpEdge(\mathit{knife}) (m_f = 0.95)\land $\\
$domain(\mathit{self},kitchen) (m_c = 1.0)\implies$ \\
\line(1,0){200}\\
$cutWith(\mathit{self},\mathit{knife}) (m_a = (m_f \otimes m_c) \odot m_r = 0.76)$
\end{flushleft}

\noindent where the $\otimes$ is the DS-theoretic uncertain logic AND
operator and the $\odot$ is the DS-theoretic uncertain logic modus
ponens operator. The uncertainty interval for the rule can then be
computed as $[0.76, 1]$.The Affordance Inference Component will then perform this analysis for
each of the other rules in the set to determine uncertainty intervals
for all the implied affordances.
Once the Affordance Inference component has found a suitable object with the required affordances, it will have completed the ``Find Object" action step and the GM will then advance the action script to the next step to ``Pick up the Object."

\textbf{Pick up Object.}
To perform this action, the GM must generate an action sequence to move near the object and then determine appropriate grasp affordances for the object in conjunction with the Affordance Inference Component. 

The Affordance Inference Component is capable of resolving not only handling functional affordances as described above with respect to finding objects to cut with, but more complex social and aesthetic affordances. For example, to properly hand over knives, it is preferable to grasp the knife by its blade and orient its handle towards a receiver. But it is acceptable to grasp the handle if the blade is dirty or being used. The Affordance Inference Component takes into account these considerations using rules of the form described above and infers appropriate grasp affordances in the current context. 

Consider the situation where the knife is dirty and the Affordance Inference Component determines that the knife is graspable by the handle. The GM selects this as a suitable grasp affordance and initiates an action sequence to execute the ``Pick up Object" action. 
Once the robot has picked up the object, it will have completed the ``Pick up Object" action step and the GM will then advance the action script to the next step to ``Bring Object to Human." 

\textbf{Bring Object to Human.}
The action of bringing an object to a human is decomposed into a simple action script that has the robot
translocating itself from its current location to the location of the human and then handing the object to
the human (handing over the object will itself be decomposed into more primitive actions). 
Given this action script, we check that the behavior of the robot is morally and socially acceptable.
These checks are made before each action is executed (including the actions described above), but for
simplicity we discuss these checks only here.
We focus on verifying that the action would be perceived as a morally acceptable behavior.  This is done
by drawing analogical comparisons with known situations and checking that similar situations are not
morally unacceptable.  If the action script to be executed next may be objectionable, then the robot tries
to modify the script and rechecks that the new script is acceptable.  Algorithm ~\ref{alg:moral} describes
this process.

\begin{algorithm}
\caption{Moral Perception Acceptability algorithm}\label{alg:moral}
\begin{algorithmic}[1]
\Procedure{CheckMoralPercept}{$s$}
	\State $m \gets$ similarScenarios($s$)
	\If{acceptable($m[0]$)}
		\Return $s$
	\Else
		\While{modifiable($s$)}
			\State $t \gets $nextModifiedActionScript($s$)
				\State CheckMoralPercept($t$)
		\EndWhile
		\State \Return error
	\EndIf
\EndProcedure
\end{algorithmic}
\end{algorithm}

The check starts with the current scenario $s$, which includes the action script to be 
executed and information about the agents and objects involved.  Given this 
description of the scenario, we find a set of \emph{similarScenarios} that analogous 
to the current one (line 2).  To compile this set of scenarios, we use SME to perform
analogical comparisons of the current scenario with other known
scenarios and return the most similar ones (up to three).  Each scenario with
which the current is compared may describe a normative action that is
taken or an action that is impermissible in the current context.  Assuming
the robot does not know of any scenario that is a \emph{literal similarity} --
such as approaching a person with a knife in a kitchen -- then we rely on
analogous scenario -- ones that are similar in structure but may differ in
content.  Consider the three scenarios described in Fig. \ref{fig:scen}.

Initially, the similarity scores of each of the scenarios is 0.4465, 0.4285, and 0.2910, respectively.  
The scores estimate the quality of the analogy but are not along any particular scale.
If the most similar scenario is morally acceptable then the current scenario does
not resemble any moral violations and the robot may proceed with executing the script (line 3).
However, our most similar scenario, the BBS, represents a moral violation.
In this case, the algorithm proceeds by considering modifications to the action script (line 6),
and then repeating the moral check on the updated scenario (line 7).
Once the action script is modified to alert the human before moving towards her, 
the HSS becomes the most analogous, and this scenario does not have any moral violations.

\begin{figure}[htpb]
\centering
\begin{mdframed}[backgroundcolor=gray!20]

\textbf{{\small Baseball Bat Scenario (BBS)}}

{\small The approaching agent surprises the approached agent
from behind and strikes them with a baseball bat. Here the approaching
agent is holding a bat, which might posses the affordances of a
weapon, and they do not provide any warning or notice to the
other. Moreover, the approaching agent is causing harm, resulting in a
morally-negative outcome.\par}

\noindent \textbf{{\small Flowers Scenario (FS)}} 

{\small One agent surprises another agent with a bouquet of
flowers.  Like the Baseball Bat scenario, here too the approaching
agent surprises the approached agent from behind, without
warning. However, unlike the Baseball Bat Scenario, here the
approaching agent is holding a bouquet of flowers, which does not
posses the affordances of a weapon. Finally, the approaching agent is
not causing harm, and in fact is cheering up the other, thereby
resulting in a morally-positive outcome.\par}

\noindent \textbf{{\small Hot Saucepan Scenario (HSS)}}

{\small One agent holding a hot saucepan warns nearby agents
while passing by behind them. Like the Flowers Scenario, here too the
outcome is a morally-positive one and the approaching agent is not
intending to cause harm. Like the Baseball Bat Scenario, here too, the
approaching agent is holding an object (hot saucepan) that possesses
weapon affordances. However, unlike both the prior scenarios, in this
scenario, the approaching agent provides a verbal warning to the
approached agent.\par}

\end{mdframed}
\caption{Analogous Scenarios}
\label{fig:scen}
\end{figure}

\section{Discussion}

Being able to recognize morally and socially charged situations is an
important skill for robots in a human-robot
collaborations. As research in robotic cognition
progresses and robots are endowed with more advanced action
capabilities, it will become ever more important to ensure that robotic
actions are monitored, discerning their moral and social implications and verifying that these actions are within societal norms. This is
especially true as robotic systems make their way into everyday
lives. Take, for example, self-driving cars. As these systems develop
the ability to monitor roads and navigate them safely, it will also be important
that they conduct themselves within social and moral expectations of
other drivers on the road. This means, therefore, looking at its own driving from
others' perspective and considering if the actions will result in
morally-positive outcomes. 


Our long-term goal is to endow robots with moral competence.  Here we
took a step in this direction by proposing promising mechanisms in an
integrated architecture for reasoning about the social and moral
propriety of situations.  Yet, many challenges remain to be addressed,
including computational complexity, episodic memory management, data
representations, as well as more advanced affordance-based and
analogical reasoning techniques.

%


%
%



\bibliographystyle{IEEEtran.bst}
\bibliography{SarathyetalHRI2016}
%
%
%

\end{document}